\pdfoutput=1

\documentclass[11pt]{article}

\usepackage[preprint]{acl}

\usepackage{times}
\usepackage{latexsym}
\usepackage{booktabs}
\usepackage{multirow}
\usepackage{tikz}
\usetikzlibrary{calc,positioning}
\usepackage{xcolor}
\usepackage{tcolorbox}
\usepackage{todonotes}

\usepackage[T1]{fontenc}

\usepackage[utf8]{inputenc}

\usepackage{microtype}

\usepackage{inconsolata}

\usepackage{graphicx}

\title{SNaRe \includegraphics[height=2ex]{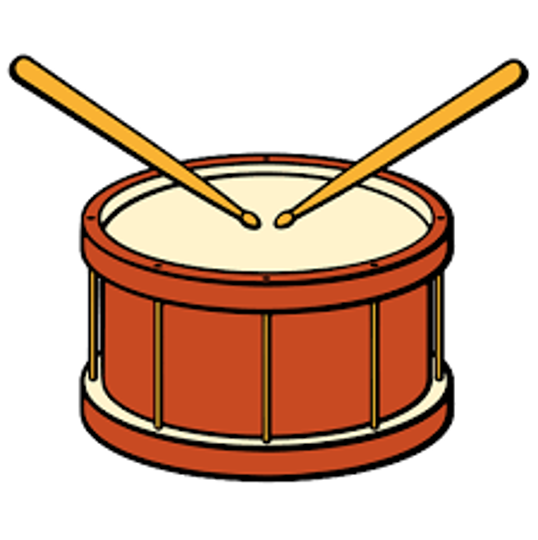}: Domain-aware Data Generation for \\ Low-Resource Event Detection}

\author{Tanmay Parekh \ \ \ \ \
Yuxuan Dong \ \ \ \ \
Lucas Bandarkar \ \ \ \ \
Artin Kim \\
{\bf I-Hung Hsu$^\dagger$ \ \ \ \ \
Kai-Wei Chang \ \ \ \ \
Nanyun Peng} \\
Computer Science Department, University of California, Los Angeles \ \ \ \ \ \ \
$^\dagger$Google \\
\texttt{\{tparekh, kwchang, violetpeng\}@cs.ucla.edu} \\
  }

\begin{document}
\maketitle

\newcommand{\mypar}[1]{\vspace{0.35em}\noindent\textbf{#1}}
\newcommand{\SideNote}[2]{\todo[color=#1,size=\small]{#2}} 

\newcommand{\tanmay}[1]{\SideNote{orange!40}{#1 --tanmay}}
\newcommand{\shawn}[1]{\SideNote{brown!40}{#1 --shawn}}

\newcommand{\cmark}{\ding{51}}%
\newcommand{\xmark}{\ding{55}}%
\newcommand{\tildemark}{\textbf{$\sim$}}%
\newcommand\splus{\ding{58}}

\newcommand{\starName}{STAR}
\newcommand{\extracttrain}{Weak Sup}
\newcommand{\modelName}{\textsc{SNaRe}}
\newcommand{\drumemoji}{\includegraphics[height=2ex]{figures/drum.png}}

\newcommand{\redtext}[1]{\textcolor{red}{#1}}
\newcommand{\bluetext}[1]{\textcolor{blue}{#1}}
\newcommand{\browntext}[1]{\textcolor{brown}{#1}}

\newcommand\blfootnote[1]{%
  \begingroup
  \renewcommand\thefootnote{}\footnote{#1}%
  \addtocounter{footnote}{-1}%
  \endgroup
}

\definecolor{triggercolor1}{RGB}{255,165,0}   
\definecolor{triggercolor2}{RGB}{30,144,255}    

\tikzset{
  event/.style={
    rectangle, draw=gray, line width=0.5pt, rounded corners,
    inner sep=2pt, align=center
  },
  event1/.style={event, fill=triggercolor2},
  event2/.style={event, fill=triggercolor1}
}

\begin{abstract}

Event Detection (ED) -- the task of identifying event mentions from natural language text -- is critical for enabling reasoning in highly specialized domains such as biomedicine, law, and epidemiology.
Data generation has proven to be effective in broadening its utility to wider applications without requiring expensive expert annotations.
However, when existing generation approaches are applied to specialized domains, they struggle with label noise, where annotations are incorrect, and domain drift, characterized by a distributional mismatch between generated sentences and the target domain.
To address these issues, we introduce \modelName{}, a domain-aware synthetic data generation framework composed of three components: Scout, Narrator, and Refiner.
Scout extracts triggers from unlabeled target domain data and curates a high-quality domain-specific trigger list using corpus-level statistics to mitigate domain drift.
Narrator, conditioned on these triggers, generates high-quality domain-aligned sentences, and Refiner identifies additional event mentions, ensuring high annotation quality.
Experimentation on three diverse domain ED datasets reveals how \modelName{} outperforms the best baseline, achieving average F1 gains of 3-7\% in the zero-shot/few-shot settings and 4-20\% F1 improvement for multilingual generation.
Analyzing the generated trigger hit rate and human evaluation substantiates \modelName{}'s stronger annotation quality and reduced domain drift.
We will release our code at \url{https://github.com/PlusLabNLP/SNaRe}.

\end{abstract}


\section{Introduction}

\begin{figure}[t]
    \centering
    \includegraphics[width=\linewidth]{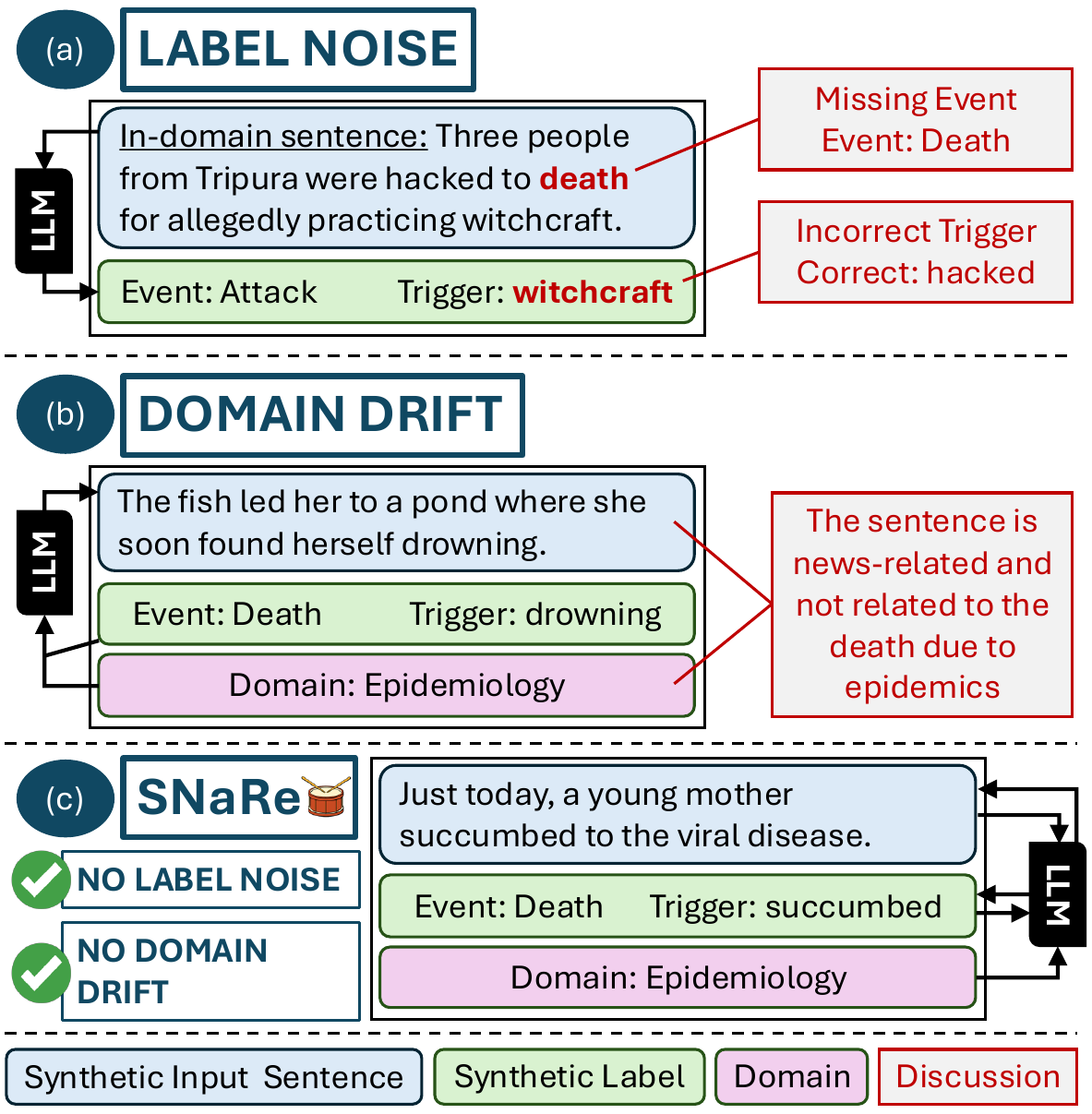}
    \caption{Highlighting the errors of existing data generation approaches. (a) Using LLMs to generate the labels from in-domain sentences leads to \textit{label noise} owing to poor LLM reasoning. (b) Utilizing LLMs to generate sentences conditioned on event and domain causes \textit{domain drift}, wherein the synthetic sentence is not aligned with the target domain. Finally, in (c), we illustrate how \modelName{} minimizes both errors to generate higher quality synthetic data.}
    \label{fig:teaser-figure}
\end{figure}

Event Detection (ED) \cite{sundheim-1992-overview, doddington-etal-2004-automatic} involves identifying and categorizing significant events from natural language text based on a pre-defined ontology. It has widespread applications in domains such as biomedicine \cite{mlee}, epidemiology \cite{parekh-etal-2024-speed, parekh-etal-2024-event}, law \cite{DBLP:conf/lrec/2010legal}.
Due to the high cost of expert-annotated data, synthetic data generation \cite{DBLP:journals/corr/abs-2312-17617} (i.e., generating sentence and event annotations) has emerged as a promising alternative, particularly for practical use-cases in specialized domains.
However, existing generation approaches often focus on general-domain settings and fail to address the distinct challenges of specialized domains \cite{llms-lack-domain-knowledge}.
Weak supervision methods \cite{DBLP:journals/corr/abs-2106-06168, chia-etal-2022-relationprompt} that use LLMs to generate labels for unlabeled sentences frequently introduce \textit{label noise} (Figure~\ref{fig:teaser-figure}(a)), where incorrect or incomplete labels arise due to weak LLM reasoning \cite{huang-etal-2024-textee} or limited domain knowledge \cite{llms-lack-domain-knowledge}.
Downstream training on such incorrect labels can cause spurious bias propagation.
Conversely, recent generation approaches \cite{josifoski-etal-2023-exploiting, star} that utilize LLMs' self-knowledge to jointly generate labels and sentences struggle with \textit{domain drift} (Figure~\ref{fig:teaser-figure}(b)), often synthesizing sentences that are misaligned with the target domain.
This can be attributed to the lack of utilization of target domain information for generation and can drastically hamper model training as lexical and structural cues are highly influential for ED \cite{tong-etal-2022-docee}.
Overall, \textit{label noise} and \textit{domain drift} reduce synthetic data quality, eventually leading to subpar supervised downstream model performance.


To this end, we propose \modelName{} \drumemoji, a novel domain-aware, three-stage data synthesis LLM pipeline comprising the \textbf{S}cout, \textbf{Na}rrator, and \textbf{Re}finer modules.
Scout surveys unlabeled target domain data to identify salient triggers via prompt-based trigger extraction.
Using corpus-level statistics for automated aggregation and filtering, Scout curates a list of high-quality domain-specific triggers per event type.
Next, Narrator samples from these domain-specific triggers and utilizes LLMs to synthesize diverse sentences for each event type.
Utilizing specialized and diverse domain information for conditional generation aids Narrator in generating more domain-aligned sentences, eventually reducing domain drift in the synthesized sentences.
Since Narrator sentences could mention additional events apart from the input set, we design the Refiner to utilize LLM inference to annotate such missing events in these sentences.
Narrator's conditional text generation and Refiner's missing label annotation aid in reducing the label noise and ensuring high data quality.
We provide an illustration of \modelName's generation in Figure~\ref{fig:teaser-figure}(c).


We benchmark \modelName{} on ED datasets from three domains: ACE \cite{doddington-etal-2004-automatic} (news), SPEED \cite{parekh-etal-2024-event} (epidemiology), and GENIA2011 \cite{kim-etal-2011-overview-genia} (biomedical).
For evaluation, we report the ED performance of DEGREE \cite{hsu-etal-2022-degree} trained on the synthesized data.
Across the zero-shot and few-shot settings, \modelName{} performs the best, outperforming the previous state-of-the-art baselines \cite{ding-etal-2023-gpt, star} by an average of 3-7\% F1 points.
Under multilingual generation for Arabic and Chinese, \modelName{} outshines even more, with improvements of 4-20\% F1 over the best baseline.
Our analysis reveals how \modelName's synthesized triggers overlap 4-11\% more (relative to baselines)  with the gold trigger set, demonstrating the reduction in domain drift.
Finally, human evaluation provides qualitative evidence for \modelName's superior data annotation quality and domain alignment.\footnote{We will release our code and data upon acceptance.}



\section{Problem Definition}

We focus on the task of Event Detection \cite{sundheim-1992-overview, doddington-etal-2004-automatic} for this work.
ED aims to extract mentions of any events of interest from natural language text.
Following ACE 2005 \cite{doddington-etal-2004-automatic}, we define an \textit{event} as something that happens or describes a change of state and is labeled by a specific \textit{event type}.
The word/phrase that most distinctly highlights the occurrence of the event is defined as the \textit{event trigger}, and the trigger-event type pair is known as the \textit{event mention}.
\textit{Event Detection} requires extracting the event \emph{triggers} from the sentence and classifying them into one of the pre-defined event types.
We provide an illustration of this task below, where \textit{arrested} and \textit{campaigns} trigger the events of \textit{Justice: Arrest-Jail} and \textit{Conflict: Demonstrate}, respectively.

\begin{tcolorbox}[width=0.48\textwidth, colback=white, colframe=gray, boxrule=0.5pt,
                  left=2mm, right=2mm, top=2mm, bottom=7mm]
Some 3,000 people have been 
\tikz[baseline, remember picture] \node[anchor=base, inner sep=0pt, outer sep=0pt] (arrested) {\textbf{\textcolor{triggercolor2}{arrested}}}; 
since the disobedience 
\tikz[baseline, remember picture] \node[anchor=base, inner sep=0pt, outer sep=0pt] (campaigns) {\textbf{\textcolor{triggercolor1}{campaigns}}}; 
began last week.
\end{tcolorbox}

\begin{tikzpicture}[remember picture, overlay]
  \node[event1] (event1) at ($(arrested.north)+(0.0,-1.20)$) 
    {\textcolor{white}{Justice: Arrest-Jail}};
  \node[event2] (event2) at ($(campaigns.north)+(-1.75,-0.725)$) 
    {\textcolor{white}{Conflict: Demonstrate}};
  
  \draw[->, thick, gray] ([yshift=-2.0ex]arrested.north) -- (event1.north);
  \draw[->, thick, gray] ([yshift=-2.0ex]campaigns.north) -- ([xshift=-3.0ex]event2.north east);
\end{tikzpicture}

In our work, we specifically focus on ED in diverse and specialized domains (e.g., biomedical), where procuring a training dataset $D_T$ of annotated data points is expensive, but unlabeled data $D_T'$ is available.
We focus on two realistic low-resource data setups - \textbf{zero-shot} (zero labeled data) and \textbf{few-shot} ($k$ labeled datapoints per event type) settings.
Unlike domain transfer, we do not consider any labeled data for the source domain, and directly optimize model performance for the target domain.


\section{Related Works}
\label{sec:related-works}

\begin{figure*}[t]
    \centering
    \includegraphics[width=0.99\linewidth]{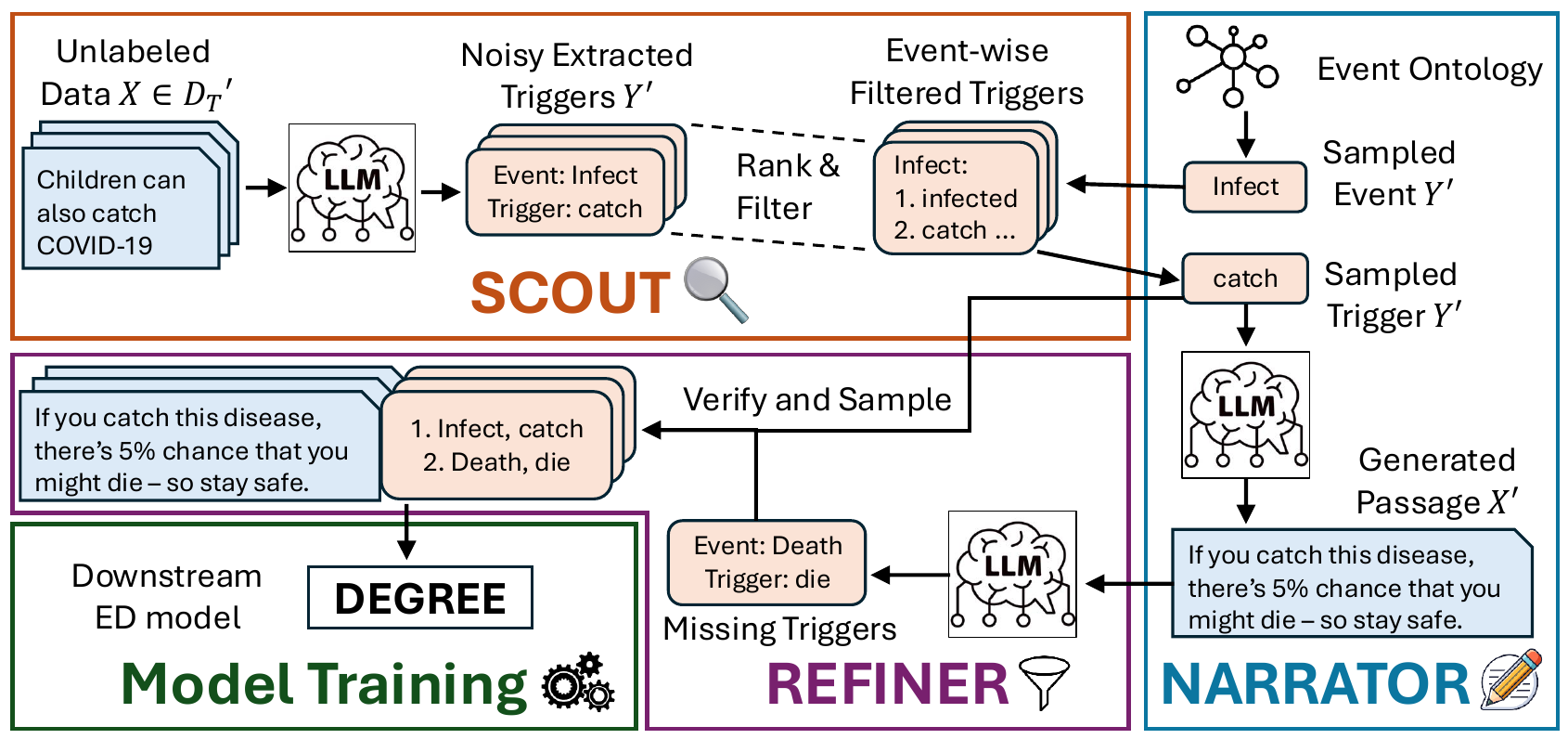}
    \caption{Model Architecture Diagram highlighting the various components of \modelName{}. First, Scout extracts and filters domain-specific triggers, then Narrator generates passages conditioned on these triggers. Finally, Refiner adds any missing annotations and sample $N$ data points per event for downstream training.}
    \label{fig:methodology}
\end{figure*}

\paragraph{Data Generation for Information Extraction}
LLM-powered synthetic data generation has been successful for various NLP tasks \cite{li-etal-2023-synthetic, wang-etal-2023-self-instruct, DBLP:journals/www/WuZQWGSQZZLXC24, shao-etal-2025-case2code}.
For information extraction, works have explored knowledge retrieval \cite{chen2023chain, amalvy-etal-2023-learning}, translation \cite{parekh-etal-2024-contextual, DBLP:conf/iclr/LeCR024}, data re-editing \cite{DBLP:journals/corr/abs-2102-01335, hu-etal-2023-entity}, and label extension \cite{zhang-etal-2024-unexpected}.
Recent works utilize LLMs to generate labels for sentences \cite{chia-etal-2022-relationprompt, ye-etal-2022-zerogen, wang-etal-2023-improving-unsupervised, DBLP:journals/corr/abs-2303-04360}, while some other works explore the generation of sentences from labels \cite{josifoski-etal-2023-exploiting, star}.
Our work introduces \modelName{} focused on infusing better domain-specific information for data generation.

\paragraph{Low-resource Event Detection}
Event Detection (ED) has been studied extensively \cite{sundheim-1992-overview, grishman-sundheim-1996-message}, leading to diverse datasets in news \cite{doddington-etal-2004-automatic, song-etal-2015-light, DBLP:conf/tac/EllisGFKSBS15}, Wikipedia \cite{li-etal-2021-document, pouran-ben-veyseh-etal-2022-mee}, and general domains \cite{wang-etal-2020-maven, parekh-etal-2023-geneva}, as well as niche areas like biomedical \cite{mlee, kim-etal-2011-overview-genia, kim-etal-2013-genia}, multimedia \cite{li-etal-2020-cross}, cybersecurity \cite{casie}, epidemiology \cite{parekh-etal-2024-speed, parekh-etal-2024-event}, and pharmacovigilance \cite{sun-etal-2022-phee}.
To address the growing need for event detection across expanding domains, prior works have explored transfer learning via Abstract Meaning Representation \cite{huang-etal-2018-zero}, Semantic Role Labeling \cite{zhang-etal-2021-zero}, and Question Answering \cite{lyu-etal-2021-zero}.
Reformulating ED as a conditional generation task has also aided low-resource training \cite{hsu-etal-2022-degree, hsu-etal-2023-ampere, huang-etal-2022-multilingual-generative}.
Recently, LLM-based reasoning \cite{DBLP:journals/corr/abs-2304-11633, DBLP:journals/corr/abs-2303-03836, wang-etal-2023-code4struct, dicore} and transfer-learning \cite{cai-etal-2024-improving-event} has been explored, but their performance remains inferior to supervised models \cite{huang-etal-2024-textee}. 
This motivates efforts in LLM-powered synthetic data generation for low-resource ED.
Although our work only demonstrates results on ED, we believe our work can be extended to other tasks outside information extraction as well, like question-answering \cite{rajpurkar-etal-2016-squad, parekh-etal-2025-dynamic} and long-form generation \cite{suvarna-etal-2024-qudselect, tartan, parekh-etal-2020-understanding}.

\section{Methodology - \modelName}
\label{sec:methodology}

In this work, we focus on domain-aware synthetic data generation using LLMs to alleviate the need for expert-annotated training data.
By generating a large dataset $D_s = \{(X,Y)\}$, we can train domain-specific downstream ED models using minimal supervised data.

Existing weak-supervision approaches \cite{mintz-etal-2009-distant, DBLP:journals/corr/abs-2109-09193} utilizing automatic methods to assign labels to unlabeled sentences often generate incorrect labels (\textit{label noise}).
This can be attributed to the domain-specific context understanding and deep reasoning requirement of ED, leading to poor automatic label quality, even when using recent LLMs \cite{huang-etal-2024-textee}.
Another route of approaches that utilize LLMs to generate sentences conditioned on labels \cite{schick-schutze-2021-generating, josifoski-etal-2023-exploiting}, i.e., synthesize $X$ for a designated $Y$, often curate sentences that are distributionally divergent from the target domain (\textit{domain drift}).
This is mainly since these approaches focus on the general domain and fail to utilize any target domain signals in their generation.
Both label noise and domain drift hurt the synthetic data quality, in turn, diminishing the downstream supervised model performance.

To mitigate these issues, we propose \modelName{} \drumemoji, a domain-aware data synthesizer, that generate and verifies LLM generations to ensure high-quality data~\cite{hsu-etal-2024-calm},
comprising three components: \textbf{S}cout, \textbf{Na}rrator, and \textbf{Re}finer:
Scout studies unlabeled target domain data $D_T'$ to curate domain-specific triggers, in turn, reducing domain drift.
Narrator generates domain-specific sentences conditioned on Scout's curated triggers, while Refiner adds additional annotations to ensure high-quality labels.
Overall, \modelName{} is a training-free LLM inference pipeline and easily deployable and scalable across domains.
We provide our architectural diagram in Figure~\ref{fig:methodology} and explain each component of our pipeline below.

\begin{figure}[t]
    \centering
    \includegraphics[width=0.95\linewidth]{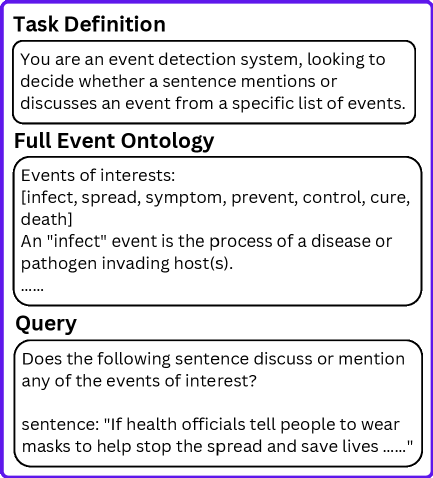}
    \caption{Prompt for stage 1 of Scout.}
    \label{fig:ed_prompt}
\end{figure}

\begin{figure}[t]
    \centering
    \includegraphics[width=0.95\linewidth]{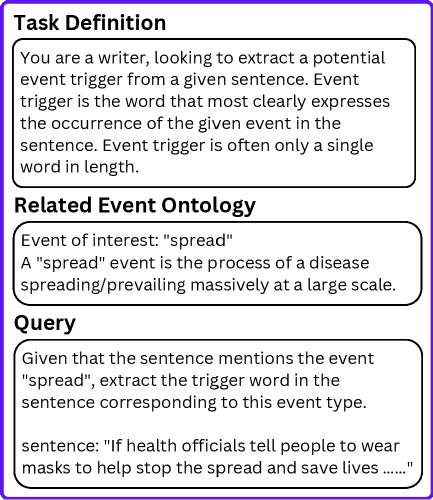}
    \caption{Prompt for stage 2 of Scout.}
    \label{fig:te_prompt}
\end{figure}

\begin{figure}[t]
    \centering
    \includegraphics[width=0.95\linewidth]{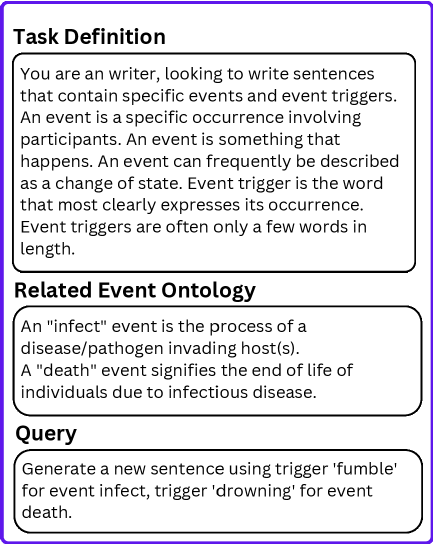}
    \caption{Prompt for Narrator.}
    \label{fig:dg_prompt}
\end{figure}

\begin{figure}[t]
    \centering
    \includegraphics[width=0.95\linewidth]{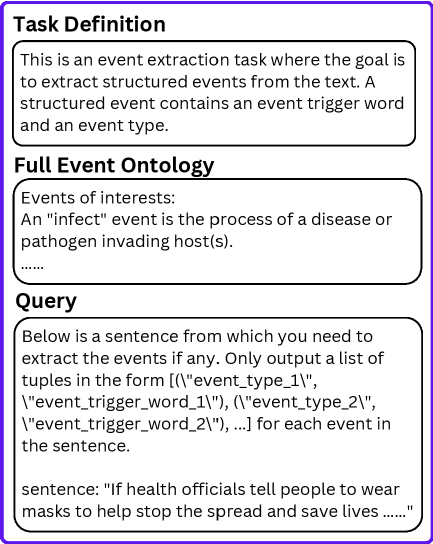}
    \caption{Prompt for Refiner.}
    \label{fig:dv_prompt}
\end{figure}

\subsection{Scout}
Scout is tasked with the curation of domain-specific triggers that are later utilized for sentence generation.
Events can assume a wide range of triggers depending on the domain and context.
For example, an "Attack" event can be triggered by \textit{war}, \textit{killed} in news, or \textit{breach}, \textit{phish} in cybersecurity, or \textit{infect}, \textit{transmit} in epidemiology domains.

Thus, unlike past works \cite{star} that utilize only LLMs' internal knowledge for trigger generation, we develop \textbf{Scout}, which extracts high-precision domain-specific triggers using unlabeled target domain data $D_T'$.
Specifically, trigger extraction involves a two-stage prompt setup:
(1) The first stage is tasked with identifying and filtering possible event types mentioned in the target domain sentence, and
(2) The second stage aims to find the most appropriate trigger word from the unlabeled sentence for each filtered event type.
We provide the prompts for Scout in Figure~\ref{fig:ed_prompt} and ~\ref{fig:te_prompt}.

To ensure high-precision of the triggers, we develop an aggregation and filtering mechanism by incorporating corpus-level statistics.
Specifically, for each event type, we aggregate the counts of the triggers at the corpus level and filter out the top $t=10$ triggers as the curated list of high-quality domain-specific event-indexed triggers.
These triggers carry important target domain signals that help in generating domain-specific sentences (\S~\ref{sec:narrator}), in turn reducing domain drift of our synthetic data.

\subsection{Narrator}
\label{sec:narrator}

Narrator is tasked with the synthesis of domain-specific sentences for our synthetic dataset.
Existing works \cite{josifoski-etal-2023-exploiting, star} do not utilize any target domain information, which causes domain drift in their synthesized sentences.
Instead, in our work, we condition the Narrator to utilize the rich and diverse domain-specific triggers from Scout to synthesize domain-specific sentences, which, in turn, reduce domain drift.

Specifically, Narrator samples 1-2 event types per synthetic data instance and corresponding domain-specific triggers from Scout's curated trigger list -- constituting the label $Y$.
Next, it prompts the LLM with the task instructions and the event definitions, and asks it to generate a passage $X'$ that mentions the sampled events using the sampled triggers ($Y$).
We illustrate this prompt in Figure~\ref{fig:dg_prompt}.
The generated sentences are naturally more aligned with the target domain owing to the conditioning on the domain-specific triggers (qualitative examples shown in \S~\ref{sec:qual-analysis}).
Further domain-specific adaptations are possible by fine-tuning the LLM on unlabeled domain-specific data (analysis in \S~\ref{sec:exgen-target-sft}).


\begin{table*}[ht]
    \centering
    \small
    \setlength{\tabcolsep}{4.5pt}
    \begin{tabular}{lll|cc|cc|cc|cc}
        \toprule
        \multirow{2}{*}{\textbf{Base LLM}} & \multirow{2}{*}{\textbf{Method}} & \textbf{Unlabeled} & \multicolumn{2}{c|}{\textbf{ACE}} & \multicolumn{2}{c|}{\textbf{SPEED}} & \multicolumn{2}{c|}{\textbf{GENIA}} & \multicolumn{2}{c}{\textbf{Average}} \\
         & & \textbf{Data Source} & \textbf{Eve-I} & \textbf{Tri-C} & \textbf{Eve-I} & \textbf{Tri-C} & \textbf{Eve-I} & \textbf{Tri-C} & \textbf{Eve-I} & \textbf{Tri-C} \\
        \midrule
        \multirow{5}{*}{Llama3-8B} & Inference & - & 30.2 & 23.8 & 39.8 & 25.4 & 21.9 & 17.2 & 30.6 & 22.1 \\
        & \starName & - & 44.9 & 35.0 & 21.0 & 10.1 & 25.9 & 19.0 & 30.6 & 21.4 \\
        & \extracttrain & train & 41.7 & 37.8 & 45.6 & 31.5 & 26.9 & 21.4 & 38.1 & 30.2 \\
        & \textbf{\modelName{} (ours)} & train & 57.4 & 50.2 & 44.6 & 31.5 & \textbf{35.2} & \textbf{28.9} & 45.7 & \textbf{36.9} \\
        & \textbf{\modelName{} (ours)} & external & \textbf{57.7} & \textbf{52.6} & \textbf{47.8} & \textbf{32.9} & 33.6 & 24.6 & \textbf{46.4} & 36.7 \\
        \midrule
        \multirow{5}{*}{Llama3-70B} & Inference & - & 46.9 & 41.3 & 46.9 & 35.6 & 34.2 & 28.2 & 42.7 & 35.0 \\
        & \starName & - & 50.0 & 42.3 & 18.3 & 13.8 & 23.3 & 16.9 & 30.5 & 24.3 \\
        & \extracttrain & train & 53.2 & 48.0 & \textbf{52.8} & \textbf{39.6} & 36.2 & 29.1 & 47.4 & 38.9 \\
        & \textbf{\modelName{} (ours)} & train & 58.1 & 53.8 & 49.9 & 38.7 & 38.0 & 29.7 & 48.7 & 40.7 \\
        & \textbf{\modelName{} (ours)} & external & \textbf{59.7} & \textbf{55.6} & 50.1 & 39.2 & \textbf{39.2} & \textbf{31.5} & \textbf{49.7} & \textbf{42.1} \\
        \midrule
        \multirow{5}{*}{GPT-3.5} & Inference & - & 33.0 & 26.2 & 44.2 & 32.9 & 31.2 & 24.7 & 36.1 & 27.9 \\
        & \starName & - & 45.0 & 36.6 & 21.3 & 14.6 & 21.8 & 14.3 & 29.4 & 21.8 \\
        & \extracttrain & train & 49.7 & 44.6 & \textbf{50.7} & \textbf{37.5} & 37.7 & 30.1 & 46.1 & 37.4 \\
        & \textbf{\modelName{} (ours)} & train & \textbf{54.8} & 48.3 & 50.3 & 36.8 & \textbf{39.3} & \textbf{31.1} & \textbf{48.1} & \textbf{38.7} \\
        & \textbf{\modelName{} (ours)} & external & 54.0 & \textbf{48.5} & 50.1 & 36.1 & 38.7 & 29.4 & 47.6 & 38.0 \\
        \midrule
        (Upper Bound) & Gold Data & - & 64.6 & 61.6 & 64.0 & 53.5 & 51.3 & 44.0 & 60.0 & 53.0 \\
        \bottomrule
    \end{tabular}
    \caption{Zero-shot results comparing \modelName{} with other baselines across three datasets and three base LLMs. Except for Inference, all other evaluations are performances of downstream DEGREE \cite{hsu-etal-2022-degree} model trained on data generated by each technique (50 datapoints per event type).}
    \label{tab:main-results}
\end{table*}

\subsection{Refiner}
While the Narrator ensures the sampled triggers are mentioned in the passage, it could potentially introduce new unknown events in the passage, leading to under-annotated $(X',Y)$ data instances.
We illustrate this in Figure~\ref{fig:missing-annotation} where the sampled trigger \textit{positive} for \textit{infect} event is mentioned, but the sentence also mentions the missing \textit{symptom} event triggered by \textit{got}.

To account for such missing annotations, we introduce \textbf{Refiner} - tasked to annotate missing events in the generated sentence.
Here, we simply prompt the LLM to find all event mentions in the generated sentence (illustrated in Figure~\ref{fig:dv_prompt}) and append them to the original sampled $Y$.
Since these new refined labels can be noisy, we avoid updating them for the already present events from the Scout, and only add them for newly discovered events.
To further improve data quality, we apply an automated rule to remove passages that do not mention the target trigger.
Additionally, we standardize trigger annotations by correcting variations in trigger word forms.
Such normalization and conservative filtering further aid in ensuring high data quality.
Finally, we apply a greedy sampling algorithm to sample $N=50$ instances $(X', Y)$ per event type to create our final synthetic dataset $D_s$.

\definecolor{eventBlue}{RGB}{41,121,255}    
\definecolor{eventRed}{RGB}{255,99,71}      
\usetikzlibrary{shapes,arrows,positioning,fit,calc,shadows}
\begin{figure}
    \centering
    \resizebox{0.85\columnwidth}{!}{
    \begin{tikzpicture}[
        base/.style={
            draw,
            rounded corners=8pt,
            inner sep=4.0pt,
            align=center,
            thick
        },
        generated/.style={
            base,
            text width=4.5cm,
            fill=eventBlue!10,
            draw=eventBlue!70,
            drop shadow={shadow xshift=3pt, shadow yshift=-3pt, opacity=0.3}
        },
        missing/.style={
            base,
            text width=2.25cm,
            fill=eventRed!10,
            draw=eventRed!70
        },
        arrow/.style={
            ->,
            >=stealth,
            line width=1.5pt,
            gray!70
        },
        arrow_missing/.style={
            arrow,
            dotted
        },
        textbox/.style={
            draw=eventBlue!70,
            dotted,
            line width=1.5pt,
            rounded corners=5pt,
            inner sep=2pt
        }
    ]
        
        \node[missing] (miss) at (4.0,0.9) {
            \textcolor{eventRed}{Missed Event Trigger:}\\[0pt]
            ``got'' \\
            \footnotesize{\textcolor{black!70}{Event: symptom}}
        };
        
        \node[generated] (gen) at (0, 1.75) {
            \textcolor{eventBlue}{Generation Trigger:}\\[0pt]
            ``Positive'' \footnotesize{\textcolor{black!70}{Event: infect}}
        };
        
        \node[align=left, text width=4.70cm] (text) at (0,0) {
            \small{Ok, I just \textcolor{eventRed}{got} a fever... Theres a possibility Im COVID-19 \textcolor{eventBlue}{Positive}}
        };
        \node[textbox, fit=(text)] {};
        
        \draw[arrow] (gen.south) -- ($(text.north)+(0.0,0.1)$);
        \draw[arrow_missing] ($(text.west)+(1.80,0.4)$) |- (miss.west);
        
    \end{tikzpicture}
    }
    \caption{Illustration of how inverse generation can produce unannotated event mentions. Blue box = target event mention, red box = unannotated event mention.}
    \label{fig:missing-annotation}
\end{figure}

\paragraph{Downstream Model Training:}
The final component utilizes the generated synthetic data $D_s$ to train downstream ED models in a supervised manner.
The trained ED models are then used to infer on the test set and for eventual evaluation.
Since we use small BART-based language models for inference, our inference time computation is negligible compared to LLM inference methods.

\section{Experimental Setup}
\label{sec:expt}

\paragraph{Datasets:}
We consider three ED datasets from diverse domains for our experiments:
(1) ACE \cite{doddington-etal-2004-automatic}, in the news domain,
(2) SPEED \cite{parekh-etal-2024-event}, in the social media domain, and
(3) GENIA \cite{kim-etal-2011-overview-genia}, in the biomedical domain. We simplify GENIA by converting the original document-level annotations to sentence-level annotations.
We consider the Arabic and Chinese versions of ACE for cross-lingual experiments.
For the few-shot setting, we sample $k$ few-shot examples from the training data.

For our unlabeled data, we consider two sources:
(1) \textbf{Train} - annotation-free training splits (i.e., only the text) of each dataset and
(2) \textbf{External} - unlabeled data from other external sources.
For this external data source, for ACE, we utilize News Category Dataset \cite{huffpost-data} comprising Huffpost news articles from 2012-2022. We filter articles corresponding to political, financial, and business articles.
For SPEED, we utilize COVIDKB \cite{zong-etal-2022-extracting}, comprising tweets from the Twitter COVID-19 Endpoint.
Finally, we utilize GENIA2013 \cite{kim-etal-2013-genia}.
We provide statistics about these datasets in Table~\ref{tab:data-statistics}.

\paragraph{Baseline methods:}
We consider three LLM-based techniques for low-resource ED as the baselines for our work.
(1) Inference \cite{DBLP:journals/corr/abs-2303-03836}: LLMs are used to directly infer on the target test data using their reasoning capability.
(2) \starName{} \cite{star}: The state-of-the-art generation model for ED utilizing LLMs for trigger and passage generation without using any unlabeled data,
(3) Weak Supervision (\extracttrain) \cite{ding-etal-2023-gpt}: LLMs are utilized to synthesize labels for unlabeled data.
For an upper bound reference, we also include a Gold Data generation baseline wherein we sample from the gold training data of each dataset to train the downstream ED model.

\paragraph{Base models:}
For our base LLMs, we consider three instruction-tuned LLMs of varying sizes, namely Llama3-8B-Instruct (8B model), Llama3-70B-Instruct (70B model) \cite{llama3}, and GPT-3.5 (175B model) \cite{gpt}.
For our downstream ED model, we consider a specialized low-resource model DEGREE \cite{hsu-etal-2022-degree}, a generative model prompted to fill event templates powered by a BART-large pre-trained language model \cite{lewis-etal-2020-bart}.

\begin{figure*}[t]
    \centering
    \includegraphics[width=\linewidth]{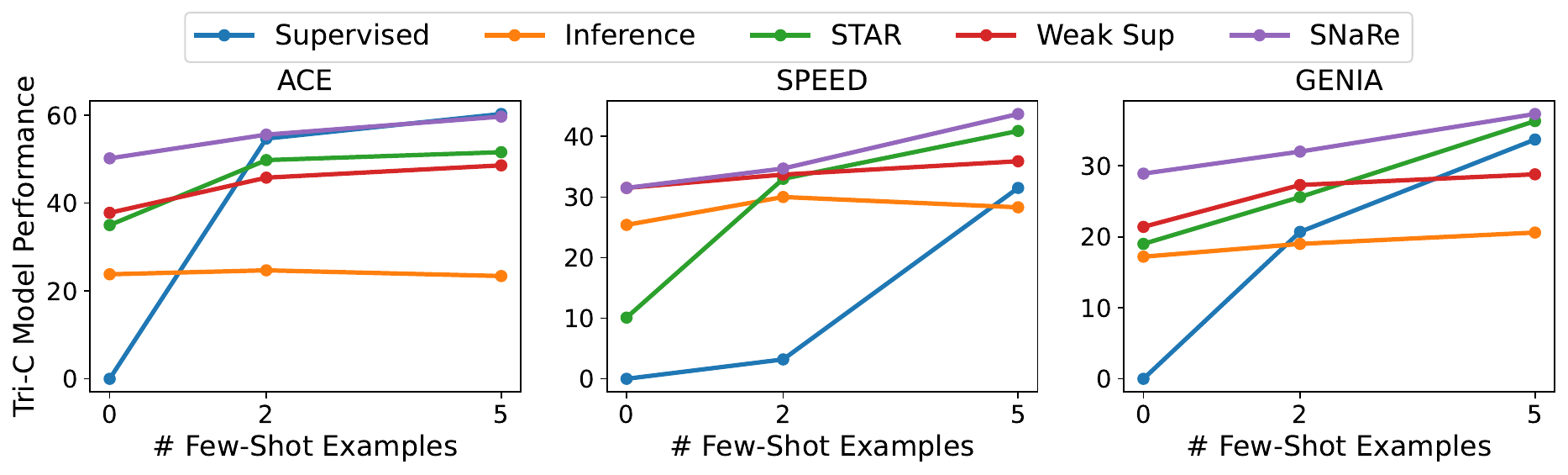}
    \caption{Few-shot results comparing \modelName{} with other baselines across three datasets using Llama3-8B-Instruct as the base LLM. Except for Inference, all other evaluations are performances of the downstream DEGREE \cite{hsu-etal-2022-degree} model trained on data generated by each technique. Tri-C: Trigger Classification F1, \#: Number of.}
    \label{fig:few-shot_tri-c_3data}
\end{figure*}

\paragraph{Evaluation:}
Our primary evaluation metric is supervised model performance trained on the synthesized data.
We consider two low-resource settings - zero-shot (no labeled data) and few-shot ($k$ datapoints per event type are used).
For Inference baseline, the LLM is directly run on the test set to procure model predictions.
We report the F1 scores for two metrics \cite{ahn-2006-stages}:
(1) Event Identification (Eve-I) - correct identification of events, and
(2) Trigger Classification (Tri-C) - correct identification of trigger-event pairs.

\paragraph{Implementation Details:}
We follow \starName{} for the implementation of the baseline models and most hyperparameter settings.
For \modelName's passage generation, we select the top $t=10$ triggers (except $t=8$ for GENIA) for passage generation.
We generate $N=50$ datapoints per event type for each generation strategy.
All our experimental results are reported over an average of three runs.
Additional details are provided in Appendix~\ref{sec:appendix-implementation-details}.

\section{Results}
\label{sec:main-results}

We present the results for our zero-shot, few-shot settings, and cross-lingual experiment below.

\subsection{Zero-shot Results}
\label{sec:zero-shot-results}

We present the main zero-shot results comparing the baselines across different LLMs and datasets in Table~\ref{tab:main-results} and discuss our findings below.


\paragraph{\modelName{} performs the best:}
On average, \modelName{} outperforms \starName{} by 17.3\% Eve-I F1 and 16.3\% Tri-C F1 points -- demonstrating how domain-specific cues from unlabeled data aid in tackling domain drift.
Compared to \extracttrain{}, \modelName{} provides average gains of 3.6\% Eve-I F1 and 3.3\% Tri-C F1, suggesting how cleaner label quality can help improve model performance.

\paragraph{External data source is effective:}
Assuming access to the training data as the unlabeled data source can be a strong assumption and bias for \modelName.
To verify the robustness of our approach, we also evaluate \modelName{} with external data sources.
Surprisingly, as seen in Table~\ref{tab:main-results}, \modelName{} with external data provides similar gains of 4\% Eve-I F1 and 3.4\% Tri-C F1 over the best baseline.
We posit that the higher volume of external data leads to the extraction of cleaner domain-specific triggers, which eventually aids in better downstream performance.

\begin{table}[t]
    \centering
    \small
    \setlength{\tabcolsep}{4.5pt}
    \begin{tabular}{ll|cc|cc}
        \toprule
        \textbf{LLM} & \textbf{Method} & \multicolumn{2}{c|}{\textbf{Arabic}} & \multicolumn{2}{c}{\textbf{Chinese}} \\
        & & \textbf{EI} & \textbf{TC} & \textbf{EI} & \textbf{TC} \\
        \midrule
        \multirow{4}{*}{Llama3-8B} & Inference & 21.5 & 13.4 & 15.0 & 11.8 \\
        & \starName{} & 11.5 & 10.5 & 19.7 & 16.0 \\
        & \extracttrain{} & 21.5 & 16.2 & 26.3 & 19.1 \\
        & \textbf{\modelName{} (ours)} & \textbf{40.1} & \textbf{33.6} & \textbf{35.9} & \textbf{31.1} \\
        \midrule
        \multirow{4}{*}{Llama3-70B} & Inference & 37.5 & 27.7 & 32.0 & 29.1 \\
        & \starName{} & 37.1 & 30.9 & 26.0 & 22.0 \\
        & \extracttrain{} & 30.0 & 20.4 & 28.1 & 26.3 \\
        & \textbf{\modelName{} (ours)} & \textbf{47.5} & \textbf{44.0} & \textbf{40.4} & \textbf{33.1} \\
        \bottomrule
    \end{tabular}
    \caption{Comparing \modelName{} with zero-shot inference for other powerful LLMs for the ACE dataset.}
    \label{tab:multilingual-results}
\end{table}

\subsection{Few-shot Results}

We also study the various methods in the presence of small annotated data as part of our few-shot experiments.
Specifically, we study the $k=2$ and $k=5$ few-shot settings, where $k$ annotated examples per event type are utilized.
We utilize the $k$-shots as in-context examples in the LLM prompts and append these few-shot examples to the synthesized training data as well.
Additionally, we consider another baseline (Supervised) of downstream models trained only on the $k$-shot examples.
We present the Tri-C results for all the datasets for the Llama3-8B model in Figure~\ref{fig:few-shot_tri-c_3data}.
Similar to zero-shot results, we observe that \modelName{} consistently outperforms all other baseline models.
On average, \modelName{} outperforms \starName{} and \extracttrain{} by 5.4\% Tri-C F1 and 7\% Tri-C F1 respectively.


\subsection{Zero-shot Multilingual Results}
\label{sec:multilingual-results}

To highlight the utility of our work, we apply our work across languages, specifically Arabic (ar) and Chinese (zh).
We used multilingual ACE data \cite{doddington-etal-2004-automatic} for this experiment and utilized TagPrime \cite{hsu-etal-2023-tagprime}, powered by XLM-Roberta-large \cite{DBLP:journals/corr/abs-1911-02116}, as the downstream ED model.
We present our results for Llama3-8B-Instruct and Llama3-70B-Instruct LLMs in Table~\ref{tab:multilingual-results}.
Surprisingly, \modelName{} performs the best out-of-the-box, with improvements ranging 10-20\% F1 for Arabic and 4-12\% F1 for Chinese,  highlighting the broader impact of our work.

\section{Analysis}

In this section, we study the superior performance of \modelName{} through various analyses.
Unless specified, we use Llama3-8B-Instruct as the base LLM.

\begin{table}[t]
    \centering
    \small
    \begin{tabular}{lccc}
        \toprule
        \textbf{Method} & \textbf{ACE} & \textbf{SPEED} & \textbf{GENIA} \\
        \midrule
        \modelName & 50.2 & 31.5 & 28.9 \\
        \quad -- Scout & 43.2 & 27.8 & 28.2 \\
        \quad -- Narrator & 37.8 & 31.5 & 21.4 \\
        \quad -- Refiner & 47.4 & 23.3 & 22.0 \\
        \bottomrule
    \end{tabular}
    \caption{Ablation study for \modelName's Scout, Narrator, and Refiner measured as Tri-C F1 performance across the three datasets.}
    \label{tab:ablation}
\end{table}

\begin{table}[t]
    \centering
    \small
    \setlength{\tabcolsep}{5pt}
    \begin{tabular}{l|cc}
        \toprule
        \textbf{LLM + Method} & \textbf{Eve-I} & \textbf{Tri-C} \\
        \midrule
        Llama3-70B + Inference & 46.9 & 41.3 \\
        Llama3.3-70B + Inference & 48.9 & 43.0 \\
        Qwen2.5-72B + Inference & 40.9 & 34.2 \\
        GPT4o-mini + Inference & 34.5 & 28.8 \\
        GPT4o + Inference & 51.4 & 47.7 \\
        QwQ-32B + Inference & 49.7 & 43.5 \\
        Deepseek-R1-L3-70B + Inference & 41.8 & 36.6 \\
        \textbf{Llama3-70B + \modelName{} (train)} & \textbf{58.1} & \textbf{53.8} \\
        \textbf{Llama3-70B + \modelName{} (external)} & \textbf{59.7} & \textbf{55.6} \\
        \bottomrule
    \end{tabular}
    \caption{Comparing \modelName{} with zero-shot inference for other powerful LLMs for the ACE dataset.}
    \label{tab:powerful-llms-results}
\end{table}

\begin{table}[t]
    \centering
    \small
    \setlength{\tabcolsep}{3.5pt}
    \begin{tabular}{l|cc|cc|cc}
        \toprule
        \textbf{Method} & \multicolumn{2}{c|}{\textbf{ACE}} & \multicolumn{2}{c|}{\textbf{SPEED}} & \multicolumn{2}{c}{\textbf{GENIA}} \\
        & \textbf{EI} & \textbf{TC} & \textbf{EI} & \textbf{TC} & \textbf{EI} & \textbf{TC} \\
        \midrule
        \modelName & \textbf{57.4} & \textbf{50.2} & \textbf{44.6} & \textbf{31.5} & \textbf{35.2} & \textbf{28.9} \\
        \extracttrain{} + \starName & 46.9 & 38.9 & 44.5 & 29.5 & 30.2 & 24.3 \\
        \bottomrule
    \end{tabular}
    \caption{Comparing \modelName{} with data-mixing of synthesized data from previous works.}
    \label{tab:results-data-mixing}
\end{table}

\begin{table}[t]
    \centering
    \small
    \begin{tabular}{lccc}
        \toprule
        \textbf{Method} & \textbf{ACE} & \textbf{SPEED} & \textbf{GENIA} \\
        \midrule
        \starName & 9.6\% & 15.3\% & 15.1\% \\
        \extracttrain & 19.1\% & 38.4\% & 44.8\% \\
        \modelName & \textbf{23.2\%} & \textbf{49.4\%} & \textbf{52.5\%} \\
        \bottomrule
    \end{tabular}
    \caption{Reporting the hit rate of synthesized data triggers relative to gold test triggers.}
    \label{tab:trigger-quality}
\end{table}

\subsection{Ablation study}
\label{sec:analysis-ablation}

Table~\ref{tab:ablation} shows the ablation study for Scout and Refiner.
For ablating Scout, we replace it by prompting LLM to directly generate triggers.
Ablating Narrator is similar to the \extracttrain baseline, while adding additional refiner and dataset statistics to remove noisy datapoints.
We observe how all the components are critical, with average performance reductions of 3.8\% F1, 6.6\% F1, and 6\% F1 upon removing the respective components of Scout, Narrator, and Refiner.


\begin{table*}[ht]
    \centering
    \small
    \setlength{\tabcolsep}{4pt}
    \begin{tabular}{l l l l p{9.8cm}}
        \toprule
        \textbf{Dataset} & \textbf{Event} & \textbf{Method} & \textbf{Trigger} & \textbf{Sentence} \\
        \midrule
        
        \multirow{4}{*}{ACE} 
          & \multirow{4}{*}{Attack} 
          & \starName  
            & raid      
            & As the rebels embarked on a daring trek across the desert, they launched a surprise \textbf{raid} on the heavily guarded fortress, catching the enemy off guard. \\
        &  & \modelName 
            & shooting  
            & As the rival businessman signed the contract, a sudden \textbf{shooting} erupted outside, causing chaos in the midst of the transaction. \\
        \midrule
        
        \multirow{4}{*}{SPEED} 
          & \multirow{4}{*}{Death} 
          & \starName  
            & asphyxiation 
            & The hiker's life was tragically cut short as \textbf{asphyxiation} occurred after she became stuck in the narrow cave crevice. \\
        &  & \modelName 
            & killed       
            & The patient's feverish state was triggered when they tested positive for the virus, which ultimately led to their being \textbf{killed} by the rapidly spreading infection. \\
        \midrule
        
        \multirow{4}{*}{GENIA} 
          & \multirow{4}{*}{Binding} 
          & \starName  
            & merge     
            & The regulatory protein's ability to activate a specific region of the DNA triggers the \textbf{merge} of two proteins, leading to the modification of gene expression. \\
        &  & \modelName 
            & bound     
            & During the phosphorylation of the enzyme, it \textbf{bound} to the DNA sequence, initiating the transcription process. \\
        
        \bottomrule
    \end{tabular}
    \caption{Qualitative examples demonstrating \starName{} and \modelName{}'s trigger and sentence generation quality.}
    \label{tab:qual-analysis}
\end{table*}

\subsection{Comparison with Powerful LLMs}
\label{sec:powerful-llms-comparison}

To further highlight the efficacy of \modelName{}, we compare it with zero-shot inference using more powerful recent LLMs for the ACE dataset in Table~\ref{tab:powerful-llms-results}.
As seen, Llama3-70B with \modelName{} significantly outperforms stronger LLMs like GPT4o and thinking-based models like QwQ-32B by 8-10\% Eve-I F1 and 8-12\% Tri-C F1 scores, respectively.
This highlights the strong efficacy of synthetic generation \modelName{} over zero-shot inference even when using more powerful LLMs.

\subsection{Comparison with Data mixing}
Data-mixing \cite{DBLP:conf/nips/HoffmannBMBCRCH22, DBLP:conf/nips/XieS0L23} is a widely used technique to leverage complementary information across datasets to promote robust downstream model training.
We mix data from our two baselines of \extracttrain{} and \starName{} as a hybrid baseline to compare with \modelName.
To keep the comparisons fair, we consider $N/2 = 25$ data instances per event type from each dataset.
Results from Table~\ref{tab:results-data-mixing} demonstrate how \modelName{} outperforms the data-mixing based hybrid model by 5-6\% F1, underlining the significance of our three-stage model design over simpler data-mixing.

\subsection{Analyzing domain drift and label quality}
\label{sec:analysis-trigger-quality}

ED models have a strong tendency to over-rely on lexical relations between triggers and events \cite{tong-etal-2022-docee}.
Thus, we compare the synthetic data triggers with the gold test triggers as a raw study of the domain drift of triggers in the synthesized data.
Specifically, we extract triggers per event type in the synthetic datasets and measure the hit rate/overlap of the synthesized triggers with the gold set of triggers, as reported in Table~\ref{tab:trigger-quality}.
\starName's low hit rate indicates the poor overlap with the gold triggers, which is a primary reason for its domain drift.
Furthermore, the consistently stronger coverage of \modelName{} explains its lower domain drift.

\begin{table}[t]
    \centering
    \small
    \begin{tabular}{lccc}
        \toprule
        \textbf{Method} & \textbf{Naturalness} & \textbf{Event} & \textbf{Annotation} \\
        & & \textbf{Relevance} & \textbf{Quality} \\
        \midrule
        \starName & 3.1 & 3.4 & 3.1 \\
        \extracttrain & \textbf{4.2} & - & 2.9 \\
        \modelName & 3.6 & \textbf{4.0} & \textbf{3.6} \\
        \bottomrule
    \end{tabular}
    \caption{Human evaluation for sentence naturalness, relevance of event in generated sentence, and the annotation quality. 1 = worst, 5 = best.}
    \label{tab:human-scoring-results}
\end{table}

\begin{table}[t]
    \centering
    \small
    \setlength{\tabcolsep}{5pt}
    \begin{tabular}{l|cc|cc|cc}
        \toprule
        \textbf{Method} & \multicolumn{2}{c|}{\textbf{ACE}} & \multicolumn{2}{c|}{\textbf{SPEED}} & \multicolumn{2}{c}{\textbf{GENIA}} \\
        & \textbf{EI} & \textbf{TC} & \textbf{EI} & \textbf{TC} & \textbf{EI} & \textbf{TC} \\
        \midrule
        \modelName & \textbf{57.4} & 50.2 & 44.6 & 31.5 & 35.2 & 28.9 \\
        \quad + SFT LLM & 55.2 & \textbf{51.7} & \textbf{46.9} & \textbf{35.8} & \textbf{36.7} & \textbf{29.1} \\
        \bottomrule
    \end{tabular}
    \caption{Measuring performance improvement by fine-tuning an LLM on unlabeled train data for \modelName.}
    \label{tab:results-fine-tuning}
\end{table}

To further study the quality and relevance of the label, we performed a human evaluation.
Specifically, a human expert in ED is tasked with scoring generations (between 1-5) on the naturalness of the sentence specific to the target domain, the relevance of the event to the semantic actions described in the generated sentence, and the annotation quality evaluating the selection of triggers (details in \S~\ref{sec:appendix-human-study}).
We provide the averaged scores across the three datasets for 90 samples in Table~\ref{tab:human-scoring-results}.\footnote{Since \extracttrain{} annotates the unlabeled target domain sentences, event relevance is analogous to annotation quality and we do not explicitly evaluate it.}
\extracttrain{} has high sentence quality but poor label annotations; \starName{} suffers from poor event relevance indicating domain drift.
Overall, \modelName{} performs the best with high annotation quality and event relevance.

\subsection{Domain-adapted LLM Fine-tuning}
\label{sec:exgen-target-sft}

We fine-tune the base LLM for Narrator on the unlabeled target domain train data $D_T'$ to better align the generated passages.
Naturally, this can be applied only for smaller LLMs owing to fine-tuning costs.
We present the results of fine-tuning Llama3-8B-Instruct on the unlabeled train data in Table~\ref{tab:results-fine-tuning}.
On average, we observe that target data fine-tuning additionally improves \modelName{} by 0.5-2\% F1.
Qualitative studies indicate that the generated passages are distributionally closer to the target domain, further reducing domain drift.


\subsection{Qualitative analysis of generated data}
\label{sec:qual-analysis}

We provide qualitative evidence for \modelName's reduction in domain drift by Scout's domain-specific triggers in Table~\ref{tab:qual-analysis} (more examples in Table~\ref{tab:complete-qual-analysis}).
We compare with \starName{}, which uses LLM's internal knowledge to generate the triggers.
Lack of domain grounding often results in \starName's triggers and sentences being misaligned (e.g. \emph{asphyxiation} for \textit{death} event related to pandemics) relative to the target domain.
In contrast, \modelName's triggers are better aligned to the target domain corpus, resulting in better quality data and reduced domain drift.





\section{Conclusion and Future Work}

We introduce \modelName{}, a domain-aware synthetic data generation approach composed of Scout, Narrator, and Refiner.
Utilizing Scout's domain-specific triggers for synthesizing sentences, along with Narrator's conditional generation and Refiner's annotations, helps reduce domain drift and label noise.
Experiments on three diverse datasets in zero-shot, few-shot, and multilingual settings demonstrate the efficacy of \modelName,
establishing \modelName{} as a strong data generation framework.

\section*{Acknowledgments}

We express our gratitude to Anh Mac, Po-Nien Kung, and Christina Chance for their valuable time, reviews of our work, and constructive feedback.
We thank the anonymous reviewers, area chairs, and the program committee for their reviews and feedback.
This work was partially supported by the National Science Foundation CAREER award \#2339766 and the Amazon AGI Research Award awarded to Nanyun Peng.

\section*{Limitations}

We consider only Event Detection (ED) as the main task for data generation, but our method can be extended to other structured prediction tasks as well.
We leave this exploration for future works.
We consider three specialized domains of news, social media, and biomedical to provide a proof-of-concept of our work.
There are other specialized domains for ED as well which can be explored as part of future work.
Finally, our proposed method \modelName{} makes a practical assumption of access to unlabeled data to procure target domain cues to guide the data generation.
However, for specific super-specialized domains or if data has privacy concerns, this may not be possible and our method may not be applicable here.
We assume such cases to be super rare and beyond the scope of our work.

\section*{Ethical Considerations}

The theme of our work is to generate high-quality domain-specific data using Large Language Models (LLMs).
The inherent LLMs can have certain biases, which can lead to potentially harmful or biased generations.
Furthermore, the LLM can introduce potential hallucinations in the annotations, which can hurt the model's performance.
We do not check or consider any bias/hallucination detection method as part of our work, as it is beyond the scope.
Future works should take due consideration of this vulnerability.

Our proposed method \modelName{} utilizes unlabeled data as a basis to procure domain-specific cues.
If there are any biases in this data, it can propagate to the downstream model as well.
We provide a proof-of-concept about our method in this work, but do not detect or rectify such biases.

\modelName's Narrator utilizes LLMs to generate sentences/passages.
However, as noticed in past work, LLMs can potentially copy these sentences from the pre-training data on which it has been trained.
This can potentially lead to copyright infringements, and we do not consider any such violations under consideration for our method.
Users should consider this vulnerability before usage in commercial applications.

We would also like to mention and acknowledge that we have utilized AI chatbots to help with the writing of the work.

\bibliography{custom,anthology}

\clearpage

\appendix





\section{Data Statistics}

We discuss details about our dataset in \S~\ref{sec:expt}.
Our test target domain data includes the test data splits of (1) ACE \cite{doddington-etal-2004-automatic} in the news domain,
(2) SPEED \cite{parekh-etal-2024-event} in the social media domain, and
(3) GENIA \cite{kim-etal-2011-overview-genia} in the biomedical domain
For unlabeled data, we utilize the training data of each dataset as one data source.
For the other data source, we utilize data from external sources, specifically:
(1) News Category Dataset (HuffPost) \cite{huffpost-data} comprising Huffpost news articles from 2012-2022 for ACE. We filter articles corresponding to political, financial, and business articles,
(2) COVIDKB \cite{zong-etal-2022-extracting} mining tweets from the Twitter COVID-19 Endpoint released in April 2020 as the external data source,
(3) GENIA2013 dataset \cite{kim-etal-2013-genia} as the external data for GENIA.
Finally, we also provide some statistics about the multilingual splits of the ACE dataset utilized for the Arabic and Chinese zero-shot experiments.\footnote{For Chinese, the average length indicates the average number of characters.}
We provide statistics about this data in Table~\ref{tab:data-statistics}.

\begin{table}[h]
    \centering
    \small
    \begin{tabular}{lccc}
        \toprule
        \textbf{Data} & \textbf{\# Sents} & \textbf{\# Event} & \textbf{Average} \\
        \textbf{Source} & & \textbf{Mentions} & \textbf{Length} \\
        \midrule
        \multicolumn{4}{c}{\textbf{Test Data}} \\
        \midrule
        ACE - test & 832 & 403 & 22.9 \\
        SPEED - test & 586 & 672 & 28.1 \\
        GENIA - test & 2,151 & 1,805 & 29.7 \\
        \midrule
        \multicolumn{4}{c}{\textbf{Unlabeled Train Data}} \\
        \midrule
        ACE - train & 17,172 & - & 15.6 \\
        SPEED - train & 1,601 & - & 33.5 \\
        GENIA - train & 6,431 & - & 30.1 \\
        \midrule
        \multicolumn{4}{c}{\textbf{Unlabeled External Data}} \\
        \midrule
        HuffPost & 43,350 & - & 17.4 \\
        COVIDKB & 7,311 & - & 30.6 \\
        GENIA2013 & 6,542 & - & 17.4 \\
        \midrule
        \multicolumn{4}{c}{\textbf{Multilingual Test Data}} \\
        \midrule
        ACE - Arabic & 313 & 198 & 24.6 \\
        ACE - Chinese & 486 & 211 & 44.2 \\
        \midrule
        \multicolumn{4}{c}{\textbf{Unlabeled Multlingual Train Data}} \\
        \midrule
        ACE - Arabic & 3,218 & - & 26.1 \\
        ACE - Chinese & 6,301 & - & 45.5 \\
        \bottomrule
    \end{tabular}
    \caption{Data Statistics for the various test and unlabeled datasets used in our work. \# = Number of.}
    \label{tab:data-statistics}
\end{table}

\section{Implementation Details}
\label{sec:appendix-implementation-details}

Here, we provide detailed implementation details for each component and the models used in our work.
We run most of our experiments on NVIDIA RTX A6000/A100 machines with support for 8 GPUs, while for GPT3.5, we make API calls through OpenAI using Curator \cite{curator}.

\subsection{LLM-based Generation}

We provide details on the various hyperparameters for using LLMs in all the components of \starName{} and \modelName.
For Llama3-8B-Instruct and Llama3-70B-Instruct, we present the hyperparameters in Table~\ref{tab:hyper-llama}; while Table~\ref{tab:hyper-gpt} presents the hyperparameters for GPT3.5.

\begin{table}[h]
    \centering
    \small
    \begin{tabular}{lr}
        \toprule
        Batch Size & 32 \\
        Temperature & 0.6 \\
        Top-p & 0.9 \\
        Max Generation Length & 250 \\
        \bottomrule
    \end{tabular}
    \caption{Hyperparameters for decoding using Llama3-8B/70B model.}
    \label{tab:hyper-llama}
\end{table}

\begin{table}[h]
    \centering
    \small
    \begin{tabular}{lr}
        \toprule
        Base LLM & gpt-3.5-turbo-0125 \\
        Temperature & 1.0 \\
        Top-p & 1.0 \\
        Max Generation Length & 500 \\
        \bottomrule
    \end{tabular}
    \caption{Hyperparameters for decoding using GPT3.5 model.}
    \label{tab:hyper-gpt}
\end{table}

\subsection{Few-shot Implementation Details}

For the few-shot setting, we can access additional $k$ datapoints per event type to aid better performance.
For LLM-based prompting, we simply add these examples in the prompt as in-context examples to help the model do better reasoning/generation.
For \starName and \modelName, we do not add the $k$ triggers to the trigger list, as it led to a drop in model performance.
This can be attributed to the presence of duplicate information, as the trigger generation/extraction already accounts for these gold triggers.
We also append the $k$ datapoints to the synthetically generated data to provide signals from the gold data.

\subsection{Downstream Model Training}

We choose DEGREE \cite{hsu-etal-2022-degree} as our downstream model for evaluation, a generation-based prompting model that utilizes natural language templates.
We implemented the DEGREE model under the TextEE framework \cite{huang-etal-2024-textee}.
Table \ref{tab:hyper-degree} presents the primary hyperparameters for this model.

\begin{table}[h]
    \centering
    \small
    \begin{tabular}{lr}
        \toprule
        Pre-trained LM & BART-Large \\
        Training Epochs & 25 \\
        Warmup Epochs & 5 \\
        Training Batch Size & 32 \\
        Eval Batch Size & 32 \\
        Learning Rate & 0.00001 \\
        Weight Decay & 0.00001 \\
        Gradient Clipping & 5 \\
        Beam Size & 1 \\
        Negative Samples & 15 \\
        Max Sequence Length & 250 \\
        Max Output Length & 20 \\
        \bottomrule
    \end{tabular}
    \caption{Hyperparameters for DEGREE model.}
    \label{tab:hyper-degree}
\end{table}

\subsection{LLM Fine-tuning}

We discuss domain-adapted passage generation through LLM fine-tuning in \S~\ref{sec:exgen-target-sft}.
Specifically, we conduct a low-rank finetuning (LoRA) \cite{DBLP:journals/corr/abs-2106-09685} to reduce computational overhead to fine-tune Llama3-8B-Instruct. We implement LoRA using the \texttt{peft} and \texttt{trl} packages \cite{peft, vonwerra2022trl}. We choose the task of causal language modeling (i.e., continual pre-training) to perform domain adaptation on unlabeled in-domain sentences. 
We utilize cross-entropy loss on the dev split of the unlabeled data to select the best model.
We provide additional details about the hyperparameters for this fine-tuning for each dataset in Table~\ref{tab:hyper-sft} below.

\begin{table}[h]
    \centering
    \small
    \begin{tabular}{lr}
        \toprule
        \multicolumn{2}{c}{\textbf{ACE}} \\
        \midrule
        Lora Rank & 32 \\
        Lora Alpha & 16 \\
        Lora Dropout & 0.1 \\
        Learning Rate & 0.0001 \\
        Weight Decay & 0.05 \\
        Training Batch Size & 32 \\
        Training Epochs & 3 \\
        Eval Steps & 20 \\
        \midrule
        \multicolumn{2}{c}{\textbf{SPEED}} \\
        \midrule
        Lora Rank & 32 \\
        Lora Alpha & 16 \\
        Lora Dropout & 0.1 \\
        Learning Rate & 0.00008 \\
        Weight Decay & 0.05 \\
        Training Batch Size & 32 \\
        Training Epochs & 10 \\
        Eval Steps & 20 \\
        \midrule
        \multicolumn{2}{c}{\textbf{GENIA}} \\
        \midrule
        Lora Rank & 32 \\
        Lora Alpha & 16 \\
        Lora Dropout & 0.1 \\
        Learning Rate & 0.00008 \\
        Weight Decay & 0.05 \\
        Training Batch Size & 32 \\
        Training Epochs & 6 \\
        Eval Steps & 20 \\
        \bottomrule
    \end{tabular}
    \caption{Hyperparameters for LoRA fine-tuning Llama3-8B-Instruct.}
    \label{tab:hyper-sft}
\end{table}

\section{Additional analyses}

In this section, we provide additional analyses to support our main experiments.




\begin{table}[t]
    \centering
    \small
    \setlength{\tabcolsep}{5pt}
    \begin{tabular}{l|cc|cc|cc}
        \toprule
        \textbf{Method} & \multicolumn{2}{c|}{\textbf{ACE}} & \multicolumn{2}{c|}{\textbf{SPEED}} & \multicolumn{2}{c}{\textbf{GENIA}} \\
        & \textbf{EI} & \textbf{TC} & \textbf{EI} & \textbf{TC} & \textbf{EI} & \textbf{TC} \\
        \midrule
        \starName & \textbf{44.9} & \textbf{35.0} & \textbf{21.0} & 10.1 & 25.9 & 19.0 \\
        \quad + mention & 44.1 & 32.9 & 17.1 & \textbf{10.3} & \textbf{28.7} & \textbf{20.4} \\
        \quad + references & 35.5 & 27.3 & 19.0 & 9.2 & 25.8 & 18.1 \\
        \bottomrule
    \end{tabular}
    \caption{Measuring model performance improvement providing domain-specific cues in the form of domain-mention (mention) or domain sentence references (references) to the LLM for \starName. EI: Event Identification F1, TC: Trigger Classification F1.}
    \label{tab:star-domain-results}
\end{table}

\begin{table}[t]
    \centering
    \small
    \setlength{\tabcolsep}{4.5pt}
    \begin{tabular}{l|cc|cc}
        \toprule
        \textbf{Method} & \multicolumn{2}{c|}{\textbf{Original ACE}} & \multicolumn{2}{c}{\textbf{Synthetic ACE}} \\
        & \textbf{EI} & \textbf{TC} & \textbf{EI} & \textbf{TC} \\
        \midrule
        LLM Training & 71.5 & 67.4 & 54.3 & 46.8 \\
        Small Model Training & 72.2 & 68.9 & 57.4 & 50.2 \\
        \bottomrule
    \end{tabular}
    \caption{Measuring model performance improvement providing domain-specific cues in the form of domain-mention (mention) or domain sentence references (references) to the LLM for \starName. EI: Event Identification F1, TC: Trigger Classification F1.}
    \label{tab:star-domain-results}
\end{table}

\begin{table}[t]
    \centering
    \small
    \setlength{\tabcolsep}{5pt}
    \begin{tabular}{l|cc|cc|cc}
        \toprule
        \textbf{Number of} & \multicolumn{2}{c|}{\textbf{ACE}} & \multicolumn{2}{c|}{\textbf{SPEED}} & \multicolumn{2}{c}{\textbf{GENIA}} \\
        \textbf{Triggers} & \textbf{EI} & \textbf{TC} & \textbf{EI} & \textbf{TC} & \textbf{EI} & \textbf{TC} \\
        \midrule
        $t=5$ & 55.9 & 48.9 & 43.2 & 29.9 & 32.2 & 26.5 \\
        $t=8$ & \textbf{57.9} & 49.7 & \textbf{45.1} & 30.8 & \textbf{35.2} & \textbf{28.9} \\
        $t=10$ & 57.4 & \textbf{50.2} & 44.6 & \textbf{31.5} & 34.8 & 28.1 \\
        $t=12$ & 56.1 & 49.3 & 44.3 & 31.2 & 33.9 & 27.4 \\
        \bottomrule
    \end{tabular}
    \caption{Ablating the impact of the number of triggers for the Scout component on downstream model performance across the three datasets. EI: Event Identification F1, TC: Trigger Classification F1.}
    \label{tab:num-triggers-ablation}
\end{table}

\begin{table}[t]
    \centering
    \small
    \begin{tabular}{p{7cm}}
        \toprule
        \multicolumn{1}{c}{\textbf{ACE}} \\
        \midrule
        A 35-year-old cyclist was hit by a speeding car while riding to work, leaving her with severe injuries, while in a separate incident, a local retail giant filed a petition to restructure its debt, sparking concerns about its financial stability. \\
        As the war on terror raged on, the Mujahideen Advisory Council distributed a statement inviting Arab and foreign media reporters to enter Fallujah and cover the battles, while simultaneously, the ownership of the ancient artifacts was transferred to the museum, with the landlord demanding rent on the premises. \\
        \midrule
        \multicolumn{1}{c}{\textbf{SPEED}} \\
        \midrule
        As the influencer's viral challenge went viral, her followers were suddenly struck with a mysterious illness after the splash of a contaminated drink, leading to a shocking explosion of fatalities on social media. \\
        As the community struggled to come to terms with the devastating accident that had claimed the lives of several residents, the authorities swiftly implemented a strict quarantine to prevent the spread of the infectious disease, hoping to mitigate the tragedy. \\
        \midrule
        \multicolumn{1}{c}{\textbf{GENIA}} \\
        \midrule
        The specific transcription factor was elevated by the presence of the hormone, thereby increasing the expression of the target gene, while the inhibitory protein curbed the activity of a competing transcription factor, preventing the expression of a repressor gene. \\
        The binding of PEBP2/CBF to the promoter region boosts the expression of the gene, which turns on the production of a crucial cytokine in response to the immune response. \\
        \bottomrule
    \end{tabular}
    \caption{Example passages of overly long and more stereotypical sentences generated when the domain is mentioned or references are added to the LLM prompt for \starName.}
    \label{tab:star-domain-examples}
\end{table}

\begin{figure*}[t]
    \centering
    \includegraphics[width=0.95\linewidth]{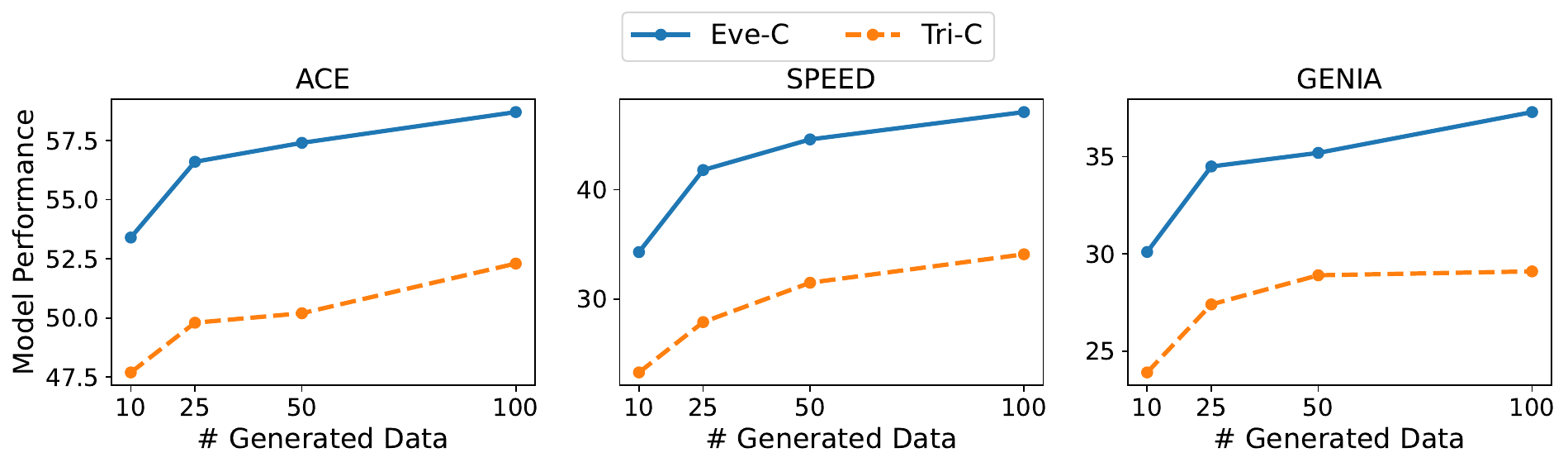}
    \caption{Model performance for \modelName{} as keep change the number of generated datapoints $N$ using Llama3-8B-Instruct for the three datasets.}
    \label{fig:appendix-numsamples-analysis}
\end{figure*}

\begin{table}[t]
    \centering
    \small
    \setlength{\tabcolsep}{5pt}
    \begin{tabular}{lcc}
        \toprule
        \textbf{Trigger Selection} & \textbf{EI} & \textbf{TC} \\
        \midrule
        Frequency Ranking (current) & \textbf{57.4} & \textbf{50.2} \\
        Sampling & 54.7 & 47.3 \\
        Weighted Sampling & 55.2 & 48.1 \\
        Reranking & 49.6 & 37.9 \\
        Minimum Count Filtering & 52.4 & 44.8 \\
        Clustering & 48.0 & 40.7 \\
        \bottomrule
    \end{tabular}
    \caption{Ablating downstream model performance on the trigger selection strategy for the Scout component on the ACE dataset. EI: Event Identification F1, TC: Trigger Classification F1.}
    \label{tab:trigger-selection-ablation}
\end{table}

\subsection{\starName{} with domain-specific prompt}
\label{sec:appendix-analysis-star-domain}

A simple way to infuse domain-specific information in past works like \starName{} would be to add domain-related information in the prompts to the LLM.
We experiment with two such methods:
(1) domain-mention, where we provide the target domain information in the prompt and ask the model to generate accordingly, and
(2) domain-reference, where we use some examples from the unlabeled data in the prompt as reference sentences to better guide the passage generation.
We provide results for these explorations using the Llama3-8B-Insturct model in Table~\ref{tab:star-domain-results}.
As observed, the results are generally poor, with an average drop of 0.1-0.6\% F1 for domain-mention and 3.1-3.8\% F1 for domain-reference.
This is mainly because LLMs over-compensate, producing longer and more stereotypical information in their generations, which hurts the naturalness of the sentence and causes further domain drift.
Furthermore, the LLM makes more errors in mentioning the event as a part of its reasoning, which is utilized to make the generation in the domain style.
We provide some qualitative examples for such generations in Table~\ref{tab:star-domain-examples}.
In some ways, it also puts into light and amplifies the gains obtained by doing target domain SFT for \modelName{} as discussed in \S~\ref{sec:exgen-target-sft}.

\subsection{Impact of different number of training samples}
\label{sec:appendix-num-samples-analysis}

We perform a small analysis to study the impact of changing the number of generated samples on the downstream model performance for \modelName.
We present the results for Llama3-8B-Instruct in Figure~\ref{fig:appendix-numsamples-analysis}.
As observed, performance continues to increase as we increase the data from $N=10$ to $N=100$ datapoints per event type.
This promises that data generation will provide continued improvements by practicing greater and better control over the data distribution.

\subsection{Training LLMs with synthetic data}
\label{sec:appendix-llm-training}

Since we generate synthetic data using LLMs, it would be natural to fine-tune LLMs to be better at Event Detection (ED). 
We conduct a small experiment to test and compare the model training for LLMs and small models.
Specifically, we train a Llama3-8B-Instruct model using Low Rank Adaptation (LoRA) \cite{DBLP:journals/corr/abs-2106-09685} and compare it with the best performing small models like DEGREE \cite{hsu-etal-2022-degree} and TagPrime \cite{hsu-etal-2023-tagprime}.
We conduct sets of training: (1) Original ACE training data, and (2) Synthetic data generated by \modelName{} for the ACE domain.
We summarize our results and findings in Table~\ref{tab:llm-training}.
Overall, we note how our synthetic data can be used to train LLMs and improve their performance iteratively as well.
However, we would like to note how small language model still learn better and provide better scores than LLMs on this task.
We would also like to note that our LLM training can be suboptimal, as we tried using the original hyperparameters and didn't tune the hyperparameters much.
We utilized Llama-Factory \cite{zheng2024llamafactory} for our LLM fine-tuning.

\subsection{Ablating the design choice for number of triggers}
\label{sec:appendix-num-triggers-ablation}

We utilize $t=10$ for ACE and SPEED and $t=8$ for GENIA.
Here, we provide additional experiments to justify this design choice.
Specifically, we provide the performance of training smaller models using different number of triggers ($t$) for the Scout using Llama3-8B-Instruct LLM in Table~\ref{tab:num-triggers-ablation}.
Overall, we observe that changing the number of triggers affects the performance slightly, and can be optimally chosen to improve performance.
However, we note that even with different $t$, \modelName{} still outperforms the other baselines.

\subsection{Ablating the design choice for trigger selection}
\label{sec:appendix-trigger-selection-ablation}

We utilize frequency-based trigger selection in the Scout, but this can lead to missing out on low-frequency, long-tail, rare triggers.
Our major motive is to avoid the noisy triggers which would also have low frequency.
In order to justify this design choice, we conduct experiments with various other trigger selection designs, specifically:
(1) Sampling: Instead of filtering and selecting, we sample uniformly from the trigger list, (2) Weighted Sampling: We sample using the extraction frequency of the triggers as the weights, (3) Reranking: Since some triggers might not be extracted correctly, we rerank the triggers based on the number of occurrences in holdout set (basically avoiding the bias of extracting them using the LLM), (4) Minimum Count Filtering: Instead of filtering based on ranks, we simply remove all triggers with a set minimum count (as this long tail might be noisy) and sample from the remaining triggers, (5) Clustering: We use k-means clustering to form clusters of similar triggers and sample one trigger from each - thus, providing enhanced diversity.
We compare these various methods for trigger selection with the Scout using Llama3-8B-Instruct LLM on the ACE dataset in Table~\ref{tab:trigger-selection-ablation}.
Overall, this study shows how the different methods, which introduce more noise and provide different ranges of the precision-recall balance, are worse in comparison to our existing frequency ranking method.

\begin{figure*}
    \centering
    \includegraphics[width=0.95\linewidth]{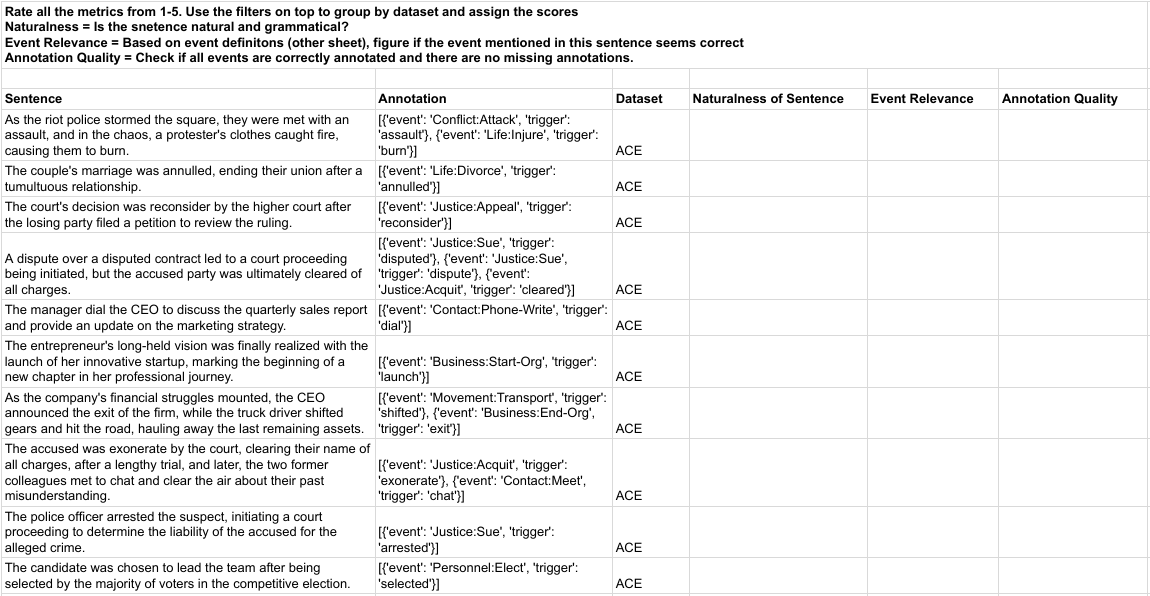}
    \caption{Illustration of the interface for the human evaluation of the synthetically generated data. Short instructions are provided at the top. Each query comprises the sentence, annotation, and dataset. The human annotator is expected to score 1-5 for each of the three metrics on the right.}
    \label{fig:human-annotation-interface}
\end{figure*}

\begin{table}[t]
    \centering
    \small
    \begin{tabular}{p{5cm}p{1cm}}
        \toprule
        \textbf{Sentence} & \textbf{Score} \\
        \midrule
        The sudden crash of the ambulance sent shockwaves through the hospital as medical staff rushed to the scene to monitor the patient's life signs, but it was too late, as the patient succumbed to the infectious disease. & SN: 2 ER: 1 AQ: 1 \\ \hline
        The wealthy entrepreneur transferred ownership of the struggling tech company to her trusted business partner, relinquishing control and financial responsibility & SN: 5 ER: 5 AQ: 5 \\ \hline
        Taken together, these data suggest that Id1 could be a possible target gene for mediating the effects of BMP-6 in human B cells, whereas Id2 and Id3 not seem to be involved. & SN: 4 ER: 3 AQ: 2\\
        \bottomrule
    \end{tabular}
    \caption{Illustration examples for the human evaluation metrics. SN: sentence naturalness, ER: event relevance, AQ: annotation quality.}
    \label{tab:human-annotation-examples}
\end{table}

\subsection{Human Evaluation Details}
\label{sec:appendix-human-study}

We conduct a small human evaluation to judge the quality of the synthetic data in \S~\ref{sec:analysis-trigger-quality}.
Here, we provide additional details about the human study and evaluation.
Since the evaluation is conducted on three diverse and niche domains, we only utilize a single human annotator who is an ED expert and has previously worked on all three datasets as the primary annotator.

We majorly evaluate on three dimensions:
(1) Sentence naturalness (SN): This metric judges whether the sentence seems grammatical, natural, and fits the domain of the target data.
(2) Event Relevance (ER): This metric is computed only for generation methods that generate sentences from labels. This evaluation judges whether the sampled event and trigger are appropriately used to generate a sensible alignment with the target domain. Furthermore, it is verified if the right event definition is used.
(3) Annotation Quality (AQ): This metric judges if the right trigger is used for each event mentioned in the synthetic output. If there are any missing events, then this score is penalized.
For each metric, a score is given on a Likert scale \cite{likert1932technique} from 1 (worst) to 5 (best).
We also provide event definitions for each event in each dataset as a reference for better judgment.
We illustrate the annotation interface in Figure~\ref{fig:human-annotation-interface} and provide some sample examples in Table~\ref{tab:human-annotation-examples}.

\begin{table}[t]
    \centering
    \small
    \setlength{\tabcolsep}{5pt}
    \begin{tabular}{l|ccc}
        \toprule
        \textbf{Method} & \textbf{\# Unlabeled Data} & \textbf{Eve-I} & \textbf{Tri-C} \\
        \midrule
        Inference & - & 46.9 & 41.3 \\
        \starName & - & 50.0 & 42.3 \\
        \extracttrain & 100\% train & 53.2 & 48.0 \\
        \extracttrain & 20\% train & 51.4 & 46.1 \\
        \extracttrain & 5\% train & 50.1 & 44.9 \\
        \midrule
        \modelName & 100\% train & 58.1 & 53.8 \\
        \modelName & 20\% train & 55.9 & 52.1 \\
        \textbf{\modelName} & 5\% train & \textbf{55.1} & \textbf{51.2} \\
        \bottomrule
    \end{tabular}
    \caption{Ablating the amount of unlabeled data utilized for the ACE dataset by the different data generation methods using Llama3-70B-Instruct and its impact on downstream model performance.}
    \label{tab:low-resource-unlabeled}
\end{table}

\subsection{Analysis in low-resource unlabeled data settings}
\label{sec:appendix-low-resource-unlabeled}

In rare cases of hyper-specialized domains or domains with privacy concerns, there is a possibility of having super-low availability of unlabeled data.
We believe that our current experimentation already simulates this setting, where we use only 773 documents for GENIA and 1.6k tweets for SPEED.
Secondly, by providing strong efficacy of external data relative to in-distribution data, we demonstrate how even mildly related unlabeled data can also work well for \modelName.
Finally, we provide a deeper study of varying the amount of unlabeled data for the ACE dataset for Llama3-70B-Instruct as an ablative study to further substantiate the efficacy of our method in Table~\ref{tab:low-resource-unlabeled}.
Lowering the number of unlabeled data reduces the model performance relative to using higher number of data samples.
However, even with 5\% unlabeled training data ($\sim$850 samples), \modelName{} still outperforms the other baselines, showing the efficacy of our method.

\begin{table}[t]
    \centering
    \small
    \setlength{\tabcolsep}{4.5pt}
    \begin{tabular}{l|cc|cc}
        \toprule
        \textbf{Method} & \multicolumn{2}{c|}{\textbf{Arabic}} & \multicolumn{2}{c}{\textbf{Chinese}} \\
        & \textbf{EI} & \textbf{TC} & \textbf{EI} & \textbf{TC} \\
        \midrule
        Inference & 33.0 & 18.7 & 25.0 & 21.0 \\
        \starName{} & 28.1 & 22.0 & 32.3 & 25.9 \\
        \extracttrain{} & 29.4 & 20.2 & 30.2 & 28.4 \\
        \textbf{\modelName{} (ours)} & \textbf{38.4} & \textbf{31.9} & \textbf{39.6} & \textbf{29.0} \\
        \bottomrule
    \end{tabular}
    \caption{Comparing \modelName{} with zero-shot inference for Qwen3-8B for the ACE dataset.}
    \label{tab:additional-zs-cross-lingual}
\end{table}

\subsection{Additional zero-shot multilingual experiments}
\label{sec:appendix-additional-zs-cross-lingual}

In \S~\ref{sec:multilingual-results}, we demonstrated the utility of our framework for zero-shot multilingual setting using the Llama3-8B and Llama3-70B models.
Here, we additionally show results using the Qwen3-8B LLM in Table~\ref{tab:additional-zs-cross-lingual}.
We note that \modelName{} outperforms the other baselines for both languages and metrics, even with the Qwen3-8B LLM.




\begin{table*}[ht]
    \centering
    \small
    \setlength{\tabcolsep}{4pt}
    \begin{tabular}{l l l l p{9.5cm}}
        \toprule
        \textbf{Dataset} & \textbf{Event} & \textbf{Method} & \textbf{Trigger} & \textbf{Sentence} \\
        \midrule
        
        \multirow{13}{*}{ACE}
            & \multirow{13}{*}{Attack}
            & \multirow{7}{*}{\starName}
                & raid
                & As the rebels embarked on a daring trek across the desert, they launched a surprise \textbf{raid} on the heavily guarded fortress, catching the enemy off guard. \\
            &
            &
                & siege
                & The government issued fines to the rebel group, while the city was under \textbf{siege}, with gunfire erupting throughout the streets. \\
            &
            &
                & slay
                & The wealthy businessman filed a complaint against his former partner, leading to a court proceeding, just as the rival gang leader \textbf{slays} his arch-nemesis in a brutal battle. \\
          \cmidrule(lr){3-5}
            &
            & \multirow{6}{*}{\modelName}
                & shooting
                & As the rival businessman signed the contract, a sudden \textbf{shooting} erupted outside, causing chaos in the midst of the transaction. \\
            &
            &
                & bombing
                & As the city was rocked by a devastating \textbf{bombing}, thousands of protesters took to the streets to demand justice and peace. \\
            &
            &
                & fight
                & The rival gangs engaged in a brutal \textbf{fight}, sparking a wave of gunfire that left several people injured and buildings damaged. \\
        \midrule

        \multirow{13}{*}{SPEED}
            & \multirow{13}{*}{Death}
            & \multirow{6}{*}{\starName}
                & asphyxiation
                & The hiker's life was tragically cut short as \textbf{asphyxiation} occurred after she became stuck in the narrow cave crevice. \\
            &
            &
                & overdose
                & A young musician's \textbf{overdose} became the tragic finale of a life cut short in the midst of chaos. \\
            &
            &
                & drowning
                & As she struggled to stay afloat, her heart stopped beating, and she succumbed to \textbf{drowning} in the icy waters. \\
          \cmidrule(lr){3-5}
            &
            & \multirow{7}{*}{\modelName}
                & killed
                & The patient's feverish state was triggered when they tested positive for the virus, which ultimately led to their being \textbf{killed} by the rapidly spreading infection. \\
            &
            &
                & lost
                & As the pandemic spread rapidly across the globe, thousands of people \textbf{lost} their lives due to the deadly virus. \\
            &
            &
                & died
                & The elderly man, who had been suffering from a severe case of tuberculosis, \textbf{died} in his sleep. \\
        \midrule
        
        \multirow{14}{*}{GENIA}
            & \multirow{14}{*}{Binding}
            & \multirow{8}{*}{\starName}
                & merge
                & The regulatory protein's ability to activate a specific region of the DNA molecule triggers the \textbf{merge} of two proteins, leading to the modification of gene expression. \\
            &
            &
                & fuse
                & When the proteins \textbf{fuse} together, the activity of the transcription factor is inhibited, preventing the gene expression from proceeding. \\
            &
            &
                & snap
                & When the two proteins \textbf{snap} together, the binding of the complex inhibits the expression of the target gene by deactivating a specific region of the DNA molecule. \\
          \cmidrule(lr){3-5}
            &
            & \multirow{6}{*}{\modelName}
                & bound
                & During the phosphorylation of the enzyme, it \textbf{bound} to the DNA sequence, initiating the transcription process. \\
            &
            &
                & translocation
                & The protein \textbf{translocation} to the nucleus triggers the induction of gene expression. \\
            &
            &
                & binds
                & When the enzyme \textbf{binds} to the substrate, it activates the addition of a phosphate group to the target molecule, marking a crucial change in its function. \\
        \bottomrule
    \end{tabular}
    \caption{Comparison of generated triggers and sentences from {\starName} and {\modelName} methods}
    \label{tab:complete-qual-analysis}
\end{table*}

\subsection{Additional Qualitative Examples}
\label{sec:appendix-qual-analysis}

In \S~\ref{sec:qual-analysis}, we discussed how \modelName{} improves domain drift qualitatively relative to \starName and provided some examples.
Here, we provide more examples to further support that study in Table~\ref{tab:complete-qual-analysis}.
This table further demonstrates how \starName{} can have a domain drift owing to a lack of domain-specific cues, while \modelName{} is better grounded in the target domain.

\end{document}